\documentclass[10pt,conference,letter]{IEEEtran}

\usepackage[english]{babel}
\usepackage[latin1]{inputenc}
\usepackage{bbm}
\usepackage{url}
\usepackage{amsmath,pifont,amssymb,amsfonts}
\usepackage{graphicx}
\usepackage{color}
\usepackage[table]{xcolor}
\usepackage{framed}
\usepackage{gensymb}
\usepackage{longtable}
\usepackage{array}
\usepackage{caption}
\usepackage{subcaption}
\usepackage{psfrag}
\usepackage{algorithm}
\usepackage{algpseudocode}
\usepackage{cite}
\usepackage{multirow}
\usepackage{balance}
\usepackage{epstopdf}
\usepackage{balance}

\graphicspath{{figs/}}

\begin{document}
\title{Two Decades of AI4NETS - AI/ML for Data Networks: Challenges \& Research Directions}

\author{
\IEEEauthorblockN{Pedro Casas}
\IEEEauthorblockA{AIT Austrian Institute of Technology, Vienna, Austria}
pedro.casas@ait.ac.at}

\maketitle

\begin{abstract}
The popularity of Artificial Intelligence (AI) -- and of Machine Learning (ML) as an approach to AI, has dramatically increased in the last few years, due to its outstanding performance in various domains, notably in image, audio, and natural language processing. In these domains, AI success-stories are boosting the applied field. When it comes to AI/ML for data communication Networks (AI4NETS), and despite the many attempts to turn networks into learning agents, the successful application of AI/ML in networking is limited. There is a strong resistance against AI/ML-based solutions, and a striking gap between the extensive academic research and the actual deployments of such AI/ML-based systems in operational environments. The truth is, there are still many unsolved complex challenges associated to the analysis of networking data through AI/ML, which hinders its acceptability and adoption in the practice. In this positioning paper I elaborate on the most important show-stoppers in AI4NETS, and present a research agenda to tackle some of these challenges, enabling a natural adoption of AI/ML for networking. In particular, I focus the future research in AI4NETS around three major pillars: (i) to make AI/ML immediately applicable in networking problems through the concepts of \emph{effective learning}, turning it into a useful and reliable way to deal with complex data-driven networking problems; (ii) to boost the adoption of AI/ML at the large scale by learning from the Internet-paradigm itself, conceiving novel distributed and hierarchical learning approaches mimicking the distributed topological principles and operation of the Internet itself; and (iii) to exploit the softwarization and distribution of networks to conceive \emph{AI/ML-defined Networks (AIDN)}, relying on the distributed generation and re-usage of knowledge through novel \emph{Knowledge Delivery Networks (KDNs)}.
\end{abstract}
\begin{IEEEkeywords}
Machine Learning; Artificial Intelligence; Data Communication Networks; Data-driven networking; Knowledge Delivery Networks (KDNs); AI/ML-defined networking (AIDN).
\end{IEEEkeywords}
\IEEEpeerreviewmaketitle

\section{Introduction}

The impressive success of AI/ML in multiple data-driven problems over the past decade has motivated a revamped interest in AI/ML-based solutions to networking problems. Data communication networks generate a wealth of big data, which offers the chance to characterize and analyze complex networking systems in a purely data-driven manner, from an end-to-end perspective, mapping input data to output targets, and without the need to model and characterize the functioning of specific components. With such wealth of big data, learning-based networking would represent a cornerstone to better, more efficient, more reliable, more flexible, and safer networks. The immediate question that poses then is: \textbf{why is networking not profiting from the outstanding success AI/ML is having in so many other data-driven domains?}. 

Making AI/ML an accepted and fruitful approach to networking in operational networks is extremely challenging. All the major learning and big data challenges - the famous \emph{4 Vs}, are present when dealing with operational networks' data: massive volumes of complex and heterogeneous data (\textbf{V}olume and \textbf{V}ariety), fast and highly dynamic streams of data (\textbf{V}elocity), lack of ground truth for learning (\textbf{V}eracity), highly unbalanced data, lack of visibility due to massive adoption of end-to-end encryption, problems for transparently interpreting AI/ML-based systems, and more. As a consequence, getting AI/ML to work in operational networks and at scale, in such a variety of dynamic data environments, with limited human intervention, and providing proper and reliable results, is still an open problem which needs to be solved. 

Despite an extensive academic research in the application of AI/ML to data communication Networks (AI4NETS) over the past two decades, there is a striking gap in terms of actual deployments of AI/ML-based solutions: compared to other network management approaches, AI/ML is rarely employed in operational scenarios.

To understand the reasons for this limited success of AI/ML in operational networks, I first present in Section \ref{secII} a brief overview on the broad domain of AI/ML, considering conventional ML and more recent developments over the past decade. I then describe past and more recent applications of AI/ML to networking problems - taking ML as the most representative approach to AI when it comes to networking. In Section \ref{secIII} I elaborate on the the major bottlenecks hindering a wide and successful adoption of AI/ML in operational networking areas. At last, in Section \ref{secIV} I present a potential research agenda in AI4NETS to advance and tackle some of these bottlenecks.

\section{State-of-the-Art}\label{secII}

\subsection{Machine Learning -- Past \& Present}

As a branch of AI or basically as a practical approach to AI, the field of ML has been studied for more than 60 years now, and today there is a plethora of ML approaches and techniques \cite{Bishop2007, Russell2010, Sutton1998, Lecun2015, Goodfellow2016}, covering the three main learning paradigms, namely supervised, unsupervised, and reinforcement learning. Very popular ML algorithms include Support Vector Machines (SVMs), Decision Trees, Neural Networks, Na\"ive Bayes, Random Forrest, Clustering approaches such as K-means and DBSCAN, Q-learning by reinforcement, and the list goes on. These more conventional ML algorithms fall within the so called \emph{shallow} learning techniques, which are basically bounded to well defined input data representations (i.e., features), and are incapable to process data in its raw form. Traditional ML has been therefore very much dependent on careful feature engineering and expert domain knowledge, to derive the most relevant set of features out of the raw data under analysis.

With the major advances in computational processing capacity, notably through the massive production and adoption of Graphics Processing Units (GPUs) and more recent Tensor Processing Units (TPUs), and the surge of data availability over the past decade, shallow learning approaches started getting overtaken by a new breed of \emph{deep} ML approaches, based on the concepts of Deep Learning (DL) \cite{Schmidhuber2015, Lecun2015, Goodfellow2016}. Different from conventional ML, DL is extremely data-driven and somehow agnostic to the specific type of data, requiring very big amounts of it to learn, and operating in a completely black-box manner. The main advantage of DL is its inherent capacity to learn directly from raw input data, without requiring heavily hand-crafted features to provide good results. The key behind DL is the so-called representation learning paradigm \cite{Bengio2013}, which offers a set of methods allowing a ML algorithm to automatically discover the best data representations or features from raw data inputs for the specific learning task (e.g., classification). DL methods are basically representation learning methods with multiple levels of representation or abstraction, obtained by composing simple but non-linear consecutive transformation steps or layers, each of them providing a more abstract representation of the data. DL has dramatically improved the state-of-the-art in multiple domains, including speech recognition, visual object recognition, object detection and many others such as genomics.

Worth mentioning are two milestones enabling a successful and massive application of DL to data-driven problems, led by the dubbed ``Godfather of DL'', Geoffrey Hinton: in 2006, Hinton et al. \cite{Hinton2006} introduced a novel and effective way to train very deep neural networks by pre-training one hidden layer at a time, using the unsupervised learning procedure for restricted Boltzmann machines \cite{Fischer2014}. In 2012, one of his students, Alex Krizhevsky, designed a deep convolutional network called the AlexNet, which strongly helped to revolutionize the field of computer vision, by almost halving the error rate for object recognition at the 2012 ImageNet challenge \cite{Krizhevsky2012}. This precipitated the rapid adoption and popularity of DL in computer vision problems, naturally extending to other domains. Also worth mentioning are the recent developments in Deep Reinforcement Learning \cite{Vincent2018}, playing a critical role in todays success of DL, notably through the popular implementations of AlphaGo, AlphaZero, and more recently AlphaStar. 

But of course, even if today everything in AI/ML is about deep learning, the AI community is extremely active with other major challenges faced by the application of ML in the practice, which together would enable the next major steps in AI. Novel learning paradigms such as the major breakthrough introduced in 2014 by Ian Goodfellow -- the Generative Adversarial Networks (GANs) approach, where models train to new levels of performance by \emph{pitting} against one other, as well as new approaches to improve the applicability of AI in the practice, by alleviating the black-box effect of complex ML models and make them more easy to understand for the end user -- eXplainable AI (XAI) \cite{Ribeiro2016,Lundberg2017}, all contribute to the popularity and successful application of ML. Other major challenges and disciplines include the problems of Continual or Lifelong Learning \cite{Parisi2018}, Robust Learning \cite{Goodfellow2018Robust}, Hierarchical Learning \cite{Bouvrie2009}, Multi-task Learning \cite{Caruana1997}, Meta Learning \cite{Lemke2015}, and Transfer Learning \cite{Dietterich1997}, among others.

\subsection{Where are we in AI4NETS?}

When it comes to the popularity of AI4NETS, the picture looks less promising. The application of AI concepts and ML approaches to networking problems has today more than two decades of existence, with the first steps taken back in the late 90's, when the concept of Cognitive networking (CN) \cite{Thomas2005} - term first coined in 1998 by researchers in KTH Sweden, was introduced. The CN paradigm describes a network with cognitive capabilities which could learn from past observations and behaviors, to better adapt to end-to-end requirements. CN has been strongly re-furbished along time, referring to it as self organizing networks, self-aware networks, self-driving networks, intelligent networks, and so forth \cite{Mahmoud2007}; this has motivated a surge of papers applying all kinds of ML algorithms to different networking problems at hand. Today we have a broad number of surveys \cite{Boutaba2018, Alsheikh2014, Bkassiny2013, Buczak2016, Fadlullah2017, Klaine2017, Nguyen2008} over-viewing the literature on the application of ML to diverse networking problems, including traffic prediction, traffic classification, traffic routing, congestion control, network resources management, network security, anomaly detection, QoS and QoE management, etc. A common trend we find in the existing literature is that, for the most part of the papers doing AI/ML for networking, there is a systematic lack of analysis on the multiple aspects which could lead to eventually re-use and reproduce, generalize, or even apply the obtained results in real deployments. In a nutshell, most of what we have in AI/ML applied to networking has been about grabbing one particular ML approach and testing it on some particular networking dataset - hopefully large, but in reality, of limited size and most probably of limited representability of the underlying problems.

\begin{figure*}[t!]
\centering
\includegraphics[width=0.85\textwidth]{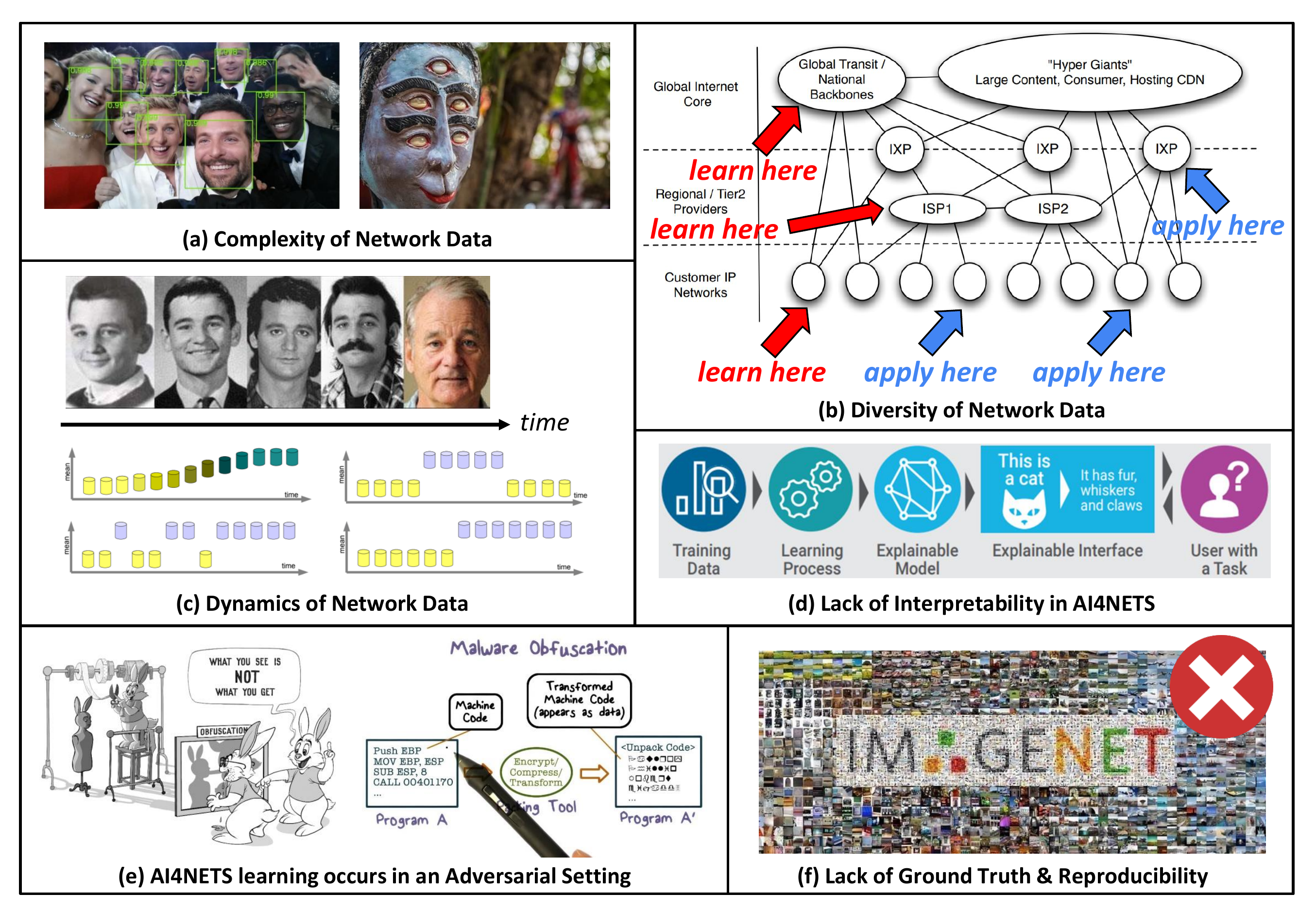}
\caption{Why is AI/ML so challenging in networking applications?}
\label{AI4NETS_limitations}
\end{figure*}

Today we start seeing papers addressing more modern flavors of AI/ML and their application in networking, e.g., with initial results on deep learning \cite{Saxe2018,Radford2018,Marin2019}, transfer learning \cite{Zhao2017}, XAI for network security \cite{Guo2018}, deep reinforcement learning for network management \cite{Nguyen2018,Mao2016,Mao2017}, etc. However, the speed of adoption of AI/ML-based solutions to current networking problems is extremely slow, with a strong resistance from not only the applied networking domain, but also the networking research community. There is a striking gap between the extensive academic research and the actual deployments of such AI/ML-based systems in operational environments. 

The picture of AI/ML in networking looks even more downcast when considering the many past efforts we have seen in turning large-scale networks into cognitive agents. Some years after the introduction of CN, Clark (one of the pioneers of the Internet) et al. \cite{Clark2003} proposed the ``Knowledge Plane'' (KP) for the Internet, a pervasive, distributed system within the network that would provide the required levels of abstraction and end-to-end visibility to realize the CN paradigm. While both CN and KP have inspired numerous research projects over the last two decades, today we still do not have any of these paradigms bringing solutions to real network deployments. With the more recent proliferation and developments on flexible networking, i.e., the Software Defined Networking (SDN) paradigm, and based on the growing availability of data coming from networking applications, there have been new attempts to realize the notions of CN and KP into operational deployments \cite{Mestres2017}, and to make of AI an enabler for data-driven networking \cite{Jiang2017}. Besides the complexities associated to SDN and network flexibility, and the limitations on network measurement and data collection in the large scale, the truth is that there are still many unsolved complex challenges associated to the analysis of networking data through ML, which hinders its acceptability and adoption in the practice. A quick thought on this already reveals part of the problem: different from other AI/ML-boosted disciplines where data comes from ``natural'' sources (image, video, voice, text) or highly constrained environments (gaming), network data lives in hard to charaterize manifolds and is highly complex, both in variety and in statistical properties (highly non-linear, statistically variant, multi-structured, etc.). Not surprisingly, already a decade ago, top networking researchers showed that the Internet has characteristics of complex systems \cite{Crowcroft2010}, making it hard to model and predict its behavior.

\textbf{The first step to unleash the power of AI in networking and to turn Clark's KP vision into a reality is to tackle the fundamental problems limiting AI's successful application in the practice}. This is exactly where I think the agenda in AI4NETS has to focus on: enabling a systematic and successful application and adoption of AI/ML in networking, by making AI/ML ``immediately applicable'' to large-scale networking problems, addressing the major bottlenecks through the application of what I have defined as \textbf{effective machine learning} and \textbf{Knowledge Delivery Networks (KDNs)}. I further elaborate on these concepts in Section \ref{secIV}. 

\section{The Challenges -- Why is AI/ML Slowly Succeeding in networking?}\label{secIII}

Let us dig deeper into the causes of such an impressive mismatch between the developments of AI/ML for networking and the popularity and success of AI in other data-driven domains. Next I provide a non-exhaustive summary on some of the main bottlenecks and challenges hindering a bright adoption of AI4NETS, including part of those previously flagged by Paxon in his seminal work on machine learning for anomaly detection \cite{Sommer2010}. I make reference to Figure \ref{AI4NETS_limitations} as a graphical guide to the different limitations we are facing in AI4NETS. 

\textbf{Data Complexity} (Figure \ref{AI4NETS_limitations}a): as I mentioned before, the complexity (and heterogeneity) of the data related to Internet-like networks is one of the most significant bottlenecks to AI4NETS. The Internet, and in general large-scale networks, are a complex tangle of networks, technologies, applications, services, devices and end-users. As a consequence, even if networking protocols are well defined (at least a-priori), the interaction among all these components makes of the resulting data a major challenge when it comes to learning anything out of it. Quoting Eric Schmidt, former Google's CEO: ``the Internet is the first thing that humanity has built that humanity does not understand, the largest experiment in anarchy that we have ever had''. Data arising from domains where AI has so far shown very successful results generaly comes from more predictable and easy to undertand sources, for example natural images, or natural language, or even from well defined and controlled environments when it comes to more recent AI victories in gaming - and not only easy Atari-like games, but more complex ones such as the recently tackled Star Craft game by Google DeepMind. For example, while one would not expect to find images of faces with five eyes and three noses when addressing face-recognition tasks, networking data can change in unpredictable manners, making it harder to deal with data-driven approaches.  

\textbf{Diversity of Network Data} (Figure \ref{AI4NETS_limitations}b): besides complexity, network data often exhibits much more diversity than one would intuitively expect. Even within a single network, the network's most basic characteristics - e.g., the mix of different applications, can exhibit immense variability, rendering them unpredictable over short time intervals. With the ever-growing number of devices connected to large-scale networks and the continuous emergence of new applications and services running on them, the diversity and unpredictability of networking data can just increase.   

\textbf{Data Dynamics} (Figure \ref{AI4NETS_limitations}c): this brings us to an even more critical issue in networking; networking data is non-static, and generally comes in the form of data streams, which by nature are difficult to analyze. Networking stream data is full of constant concept drifts - changes in the underlying statistical properties, which requires different from traditional approaches to make sense and good use out of it \cite{Mulinka2018}. 

\textbf{Lack of Learning Generalization} (Figure \ref{AI4NETS_limitations}b): as a consequence, it becomes extremely difficult in the networking practice to learn models which can generalize to other environments, different from those where the training data comes from. A very common issue in learning for networking problems is that those solutions which are built and calibrated for certain types of networks, are no longer useful when deployed in the operational environments.

\textbf{Lack of Ground Truth} (Figure \ref{AI4NETS_limitations}f): supervised learning needs massive amounts of labeled data to learn, but ``in the wild'' networking data is usually non-labeled; labeling data in operational environments is extremely costly and error prone. As a result, AI/ML-based solutions have to heavily rely on unsupervised or semi-supervised learning schemes, with their corresponding limitations in terms of knowledge generation.

\textbf{Dealing with highly Imbalanced Problems}: network data is usually highly imbalanced when it comes to classes representing different categories in the data; for example, network anomalies and attacks reasonably occur much less often than normal instances; this has to be properly taken into account to avoid over-fitting and other undesirable learning problems.

\textbf{Lack of Standardized and Representative Datasets} (Figure \ref{AI4NETS_limitations}f): different from other AI/ML-related domains, where well established, publicly available datasets are available for testing, evaluation and benchmarking purposes (e.g., ImageNet in image processing), it is very difficult to find appropriate public datasets to assess AI4NETS. While one of the main reasons for this lack clearly arises from the data's sensitive nature - including end-user privacy, other limitations come from the efforts required to build proper and representative datasets in networking. Given the scale of Internet-like networks, the massive volumes of data, and the multiplicity of operational conditions, building such a representative dataset is a daunting task. So in simple words, there is no ImageNet in AI4NETS. 

\textbf{Lack of Model Performance Bounds \& High Cost of Errors}: deploying AI/ML-based solutions in operational environments comes with the associated challenge of incurring in costly errors - specially when dealing with critical applications such as security, which usually ISPs and network vendors are not willing to bear. Being data-driven by nature, and prone to outliers, it is extremely challenging to provide tight performance bounds on trained AI models. Robust learning is paramount for networking. 

\textbf{Learning occurs in an Adversarial Setting} (Figure \ref{AI4NETS_limitations}e): learning data is full of adversarial examples in many networking problems, which again makes the learning process more cumbersome; this is the case not only in network security applications, where a classic arms-race is evident, but also when it comes to traffic identification and classification tasks, as applications do not want to be tracked by intermediate entities, and therefore obfuscate and dynamically modify their functioning to bypass monitoring and avoid traffic engineering policies.

\textbf{Lack of Model Transparency} (Figure \ref{AI4NETS_limitations}d): the lack of transparency of most ML models limits their application in real deployments, especially in critical applications such as security. If you cannot understand why this or that decision is taken, then you would not trust it and therefore not use it. The complexity of current DL models exacerbates this black-box effect, therefore limiting its application in the practice. 

\textbf{Lack of a combined Knowledge \& Expertise in both networking and AI/ML}: the networking and the AI/ML communities are totally separated, and it is very hard today to find the required expertise to properly tackle networking problems through AI/ML. So far, AI/ML in networking has been about grabbing one particular ML approach and testing it on (hopefully) large-scale datasets, without really mastering the underlying properties and requirements of the selected ML approach. At the same time, we have seen AI researchers applying ML to networking problems which they do not really understand, and therefore the validity of the obtained conclusions and their analysis is quite limited in the best case. There is a huge need of interdisciplinary approaches and teams to tackle this limitation.

\textbf{Up to date, there has not been any systematic research effort towards better understanding and tackling the main issues limiting a successful adoption of AI to data-driven networking problems.} As a consequence, the lack of success-stories in AI4NETS is not surprising at all.

\section{A Research Agenda in AI4NETS}\label{secIV}

\begin{figure}[t!]
\centering
\includegraphics[width=0.85\columnwidth]{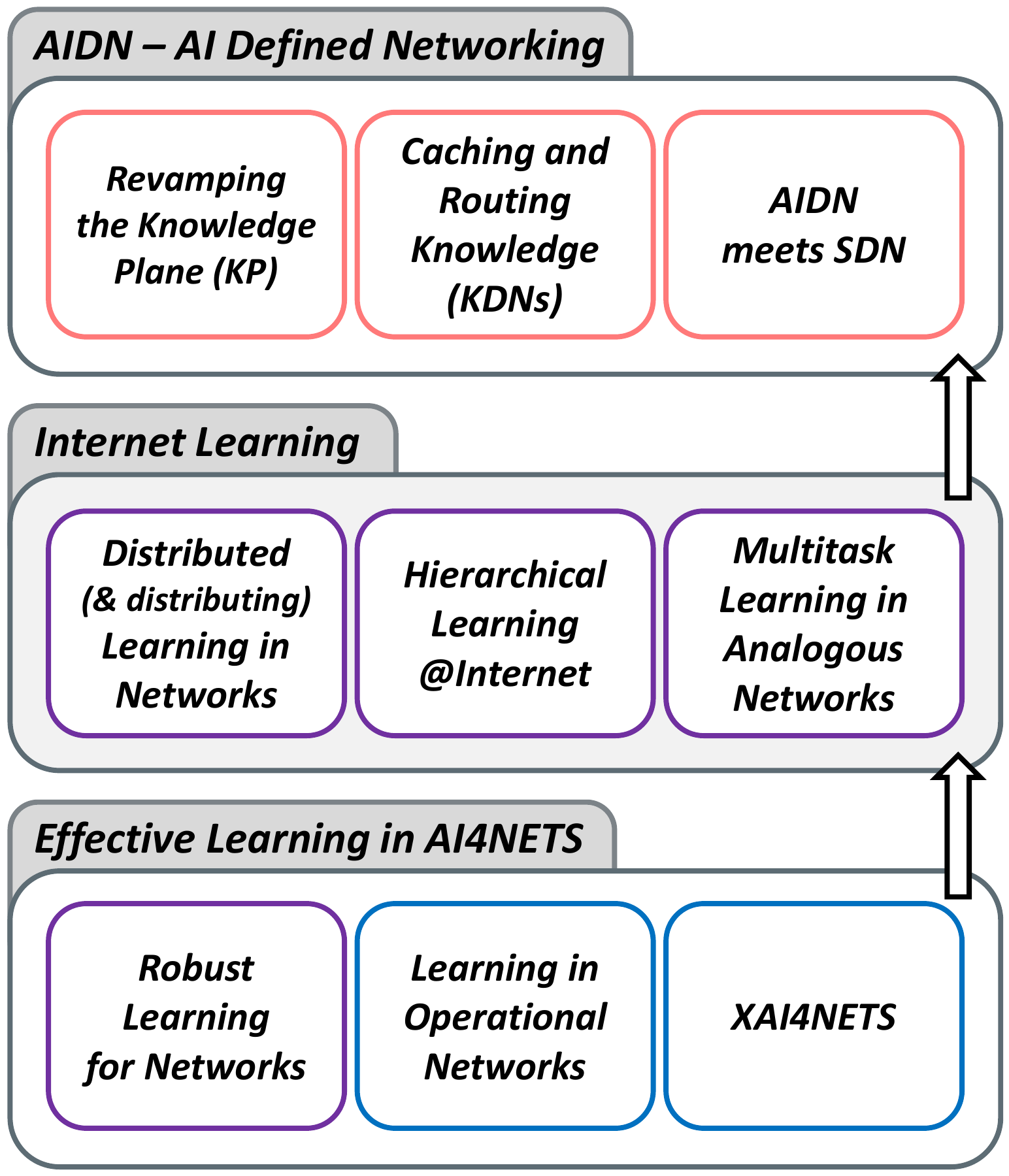}
\caption{A research agenda in AI4NETS.}
\label{AI4NETS_agenda}
\end{figure}

In the next lines I propose an ambitious research agenda to materialize novel ideas, concepts, and learning approaches towards \textbf{bridging the gap between AI/ML and its successful and systematic application to improve networking problems at the large scale}. The so far presented state of affairs in AI4NETS clearly suggests that resolving the main bottlenecks hindering a natural adoption of AI/ML in data-driven networking problems requires a solid grasp on the main concepts and principles behind network measurements and AI/ML. In Figure \ref{AI4NETS_agenda}, I depict the three main pillars of the proposed research agenda, including (i) effective learning, (ii) Internet (distributed) learning, and (iii) AI defined Networks.

\textbf{Effective Learning in AI4NETS} deals with making AI/ML immediately applicable to networking problems. By immediately applicable, I mean a successful application in terms of performance, generalization of results, and model transparency, leading to a subsequent natural adoption in networking problems. The principles behind what I shall refer to as \textbf{effective learning} for networks consists of a series of general AI/ML-related problems (robust, transferable and explainable learning), applied to the particular networking challenges outlined above. The work I propose in this area can be further structured under three specific topics -- \textbf{(i) Robust learning for networking:} consists of the investigation, tailoring and application of more novel learning paradigms which are currently showing great performance in other domains, including (among others) robust-deep-learning based paradigms and representation learning, (deep) reinforcement leaning for closed-loop applications, and the study of robust transfer in networking learning tasks. All these approaches have evolved driven by non-networking applications (e.g, image processing), and as such, need to be seriously re-considered when dealing with networking data. \textbf{(ii) Learning in operational networks:} should tackle those challenges which arise from the networking practice, when dealing with real networks data and when deploying AI/ML-based solutions into operational settings; learning with highly imbalanced data, learning with limited (or none) availability of labels/ground truth, or learning in highly dynamic settings and under the presence of frequent concept drifts are some of the problems tackled here. \textbf{(iii) Explainable AI for networks:} addresses one of the most important challenges stopping the adoption of AI/ML-based solutions, namely the transparency and understanding of the decisions taken by ML models in networking problems. Conceiving approaches to better explain model properties, explain the specific rationale behind decisions, and validate the operation of black-box like models, will ease their adoption in operational environments.

\textbf{Internet Learning} is complementary to the first pillar, but instead of asking \emph{what can AI/ML do for the Internet}, the target here is to figure out \textbf{\emph{what can the Internet do for AI/ML}}; in particular, here I propose a radically different approach to design distributed learning algorithms, by leveraging the underlying properties of the Internet itself in terms of distributed topological principles and operation, to build better and more robust learning approaches for networking. As noted by Stonebraker - one of the gurus of big data analytics, distributed learning is the next frontier to AI/ML scalability \cite{Brodie2018}. One major bottleneck for scalable and distributed AI/ML is that most ML algorithms were not originally conceived to operate in a distributed manner, which seriously limits the availability of ML libraries for parallel computing platforms. To realize this Internet learning concept, and as noted in \cite{Blenk2017}, I propose to leverage the fact that distributed network algorithms generate a wealth of big data in the form of problem instance and solution pairs as part of their normal operation, which introduces an unexplored opportunity to learn from this examples at a massive scale. The work I propose in this area can be further structured under three specific topics -- \textbf{(i) Distributed (and distributing) learning in networks:} here the focus is on the principles behind distributed machine learning and distributed computation \cite{Dean2012,Crotty2014,Zamanian2017}, conceiving approaches to realize such distributed learning by relying on distributed network measurements. \textbf{(ii) Hierarchical learning at the Internet:} investigates the main properties of hierarchical learning approaches, through the application of divide \& conquer paradigms to solve complex learning tasks, breaking them up into smaller components. The Internet itself has a highly hierarchical structure, thus the synergies and parallels between learning with hierarchical data and the underlying hierarchical networks is an appealing topic to explore. \textbf{(iii) Multi-task learning:} lastly, I put the emphasis on learning from multiple, heterogeneous networking data, aiming to generate knowledge from the large amount of data available in large scale networks. I propose to leverage the data contained in multiple related tasks to improve overall learning generalization, by solving multiple learning tasks at the same time, while exploiting commonalities across tasks.

Finally, \textbf{AI Defined Networking} has as main objective to make of AI/ML and the knowledge distributively generated at the Internet scale a driving force for flexible network management and operation. With the recent developments and rising deployment of SDN at the large scale \cite{SDN2015}, there is now a bright new path to couple this flexibility with the knowledge generated by AI/ML. Here I propose to \textbf{(i) revamp the KP (Knowledge Plane)} concepts \cite{Clark2003}, focusing in particular on the KP architecture in terms of availability and accessibility of the knowledge (trained-models, building-blocks) generated across multiple networks. I propose to investigate different concepts and architectural designs to make this knowledge flowing through the Internet, the same way routing information and content propagate in current networks. How to store and distribute such knowledge brings me to a novel proposal, that of \textbf{(ii) Knowledge Delivery Networks (KDNs)}. A KDN replicates the same concepts of current Content Delivery Networks (CDNs), but takes into account the potential and characteristics of the particular content, similarly to the Information-Centric Networking (ICN) paradigm \cite{Koponen2007}; using the techniques and results obtained through effective, Internet learning, a KDN could build, maintain and merge different pieces of \emph{knowledge-blocks} to improve learning and replicate lessons learned. The KDN concept would eventually lead to a knowledge transfer system, where different models learned in specific environments and conditions could be re-used, to improve new learning steps. Similar ideas have been recently proposed \cite{Mehari2018}, relying on Cloud repositories. A KDN overlay network could be initially realized through well established CDN networks. \textbf{(iii) AIDN meets SDN:} the ultimate integration of the concepts behind Internet Learning, the foundations of the KP, and the realization of KDN networks withing a SDN architecture, would lead to the next generation of AI/ML-defined (driven and capable) networks.

\section{Conclusions}

The ultimate goal of this positioning paper is to contribute to strengthening the future research on the (re)-emergent field of AI4NETS, by pinpointing some of the fundamental challenges it faces. I acknowledge that the discussion I present in this paper is by no means fully exhaustive and final, and I am sure many aspects were left aside; however, I do hope it will motivate further discussion on the limitations we are facing in AI4NETS as a research community, to better target the next wave of AI/ML-based solutions in the practice. 

\balance

\bibliographystyle{nature}
\bibliography{biblio} 
\end{document}